\documentclass{article}
\usepackage{ijcai20}

\usepackage{times}
\usepackage{soul}
\usepackage{url}
\usepackage{xcolor}
\usepackage[hidelinks]{hyperref}
\usepackage[utf8]{inputenc}
\usepackage[small]{caption}
\usepackage{graphicx}
\usepackage{amsmath}
\usepackage{amsthm}
\usepackage{booktabs}
\usepackage{algorithm}
\usepackage{algorithmic}
\title{TransSent: Towards Generation of Structured Sentences with Discourse Marker}

\author{
Xing Wu$^3$\footnote{Work done while the author was a master at University of Chinese Academy of Sciences.}\footnotemark[4] \and
Dongjun Wei$^{1,2}$\footnotemark[4] \and
Liangjun Zang$^{1}$\and
Jizhong Han$^{1}$\And
Songlin Hu$^{1,2}$\\
\affiliations
$^1$Institute of Information Engineering, Chinese Academy of Sciences, Beijing, China\\
$^2$School of Cyber Security, University of Chinese Academy of Sciences, Beijing, China\\
$^3$Baidu Inc., Beijing, China
\emails
wuxing03@baidu.com,
\{weidongjun, zangliangjun, hanjinzhong, husonglin\}@iie.ac.cn
}

\begin{document}

\maketitle
\footnotetext[4]{Equally contributed.}
\begin{abstract}
Structured sentences are important expressions in human writings and dialogues.
Previous works on neural text generation fused semantic and structural information by encoding the entire sentence into a mixed hidden representation. 
However, when a generated sentence becomes complicated, the structure is difficult to be properly maintained.
To alleviate this problem, we explicitly separate the modeling process of semantic and structural information.
Intuitively, humans generate structured sentences by directly connecting discourses with discourse markers (such as \emph{and}, \emph{but}, etc.). Therefore, we propose a task that mimics this process, called \textbf{discourse transfer}.
This task represents a structured sentence as $($\emph{head discourse, discourse marker, tail discourse}$)$, and aims at tail discourse generation based on head discourse and discourse marker.
We also propose a corresponding model called \textbf{TransSent}, which interprets the relationship between two discourses as a translation\footnote{In this paper, we use word ``transfer" to represent a general process of our task and word ``translation" to represent the operation in embedding space. In addition, we use the phrases ``embedding space" and ``hidden representation space" interchangeably.} from the head discourse to the tail discourse in the embedding space.
We experiment TransSent not only in discourse transfer task but also in free text generation and dialogue generation tasks. 
Automatic and human evaluation results show that TransSent can generate structured sentences with high quality, and has certain scalability in different tasks.
\end{abstract}

\section{Introduction}
Automatically generating well-structured and semantically meaningful text has various applications in question answering, dialogue systems, product reviews, etc. 
In recent years, with the development of deep neural networks, many studies have produced encouraging results in natural language generation.
Some generate sentences from scratch~\cite{bowman2015generating,yu2017seqgan,Sutskever-2014-NIPS,rajeswar2017adversarial}, 
while others~\cite{kikuchi2016controlling,hu2017toward} explore the influence of different attributes in sentences, such as \emph{positive} and \emph{negative}. There are also some works leverage discourse-level information to improve the generation quality~\cite{xiong2018modeling,zhao2017learning}.
These models use an end-to-end encoder-decoder network. The encoder fuses semantic and structural information into a mixed hidden representation, and the decoder decodes the representation into a sentence.
However, when a generated sentence becomes complicated, the structure is difficult to properly maintain. Although there are some works on long or diverse text generation~\cite{guo2018long,bosselut2018discourse,shao2019long,shen2019towards}, these models mainly focus on generating cohesive or coherent text, rather than explicit sentence structures.

\begin{table}[t]
  \small
    \centering
    \caption{Examples of discourse pairs with correct discourse markers.} \label{discourse pairs}
    \label{tab1}
    \begin{tabular}{c|c|c}
    \toprule
    S1& marker& S2\\
    \midrule
    She was late to class& because& she missed the bus\\
    She was good at soccer& but& she missed the goal\\
    She had a clever son& and& she loved him\\
    \bottomrule
    \end{tabular}
\end{table}

Intuitively, humans generate structured sentences by directly connecting discourses with discourse markers~\footnote{Discourse markers are the words that mark the semantic relationship between two discourses, such as \textit{because, but, and}.}~\cite{hobbs1990literature}, as shown in Table~\ref{discourse pairs}. In neural sentence generation, we can also try to use discourse markers to constrain the structure of a sentence. Therefore, the model can focus on the modeling process of semantic information. This is much easier than generating an entire structured sentence from a mixed hidden representation.

\begin{figure}[t]
	\begin{centering}
	\includegraphics[width=0.4\textwidth]{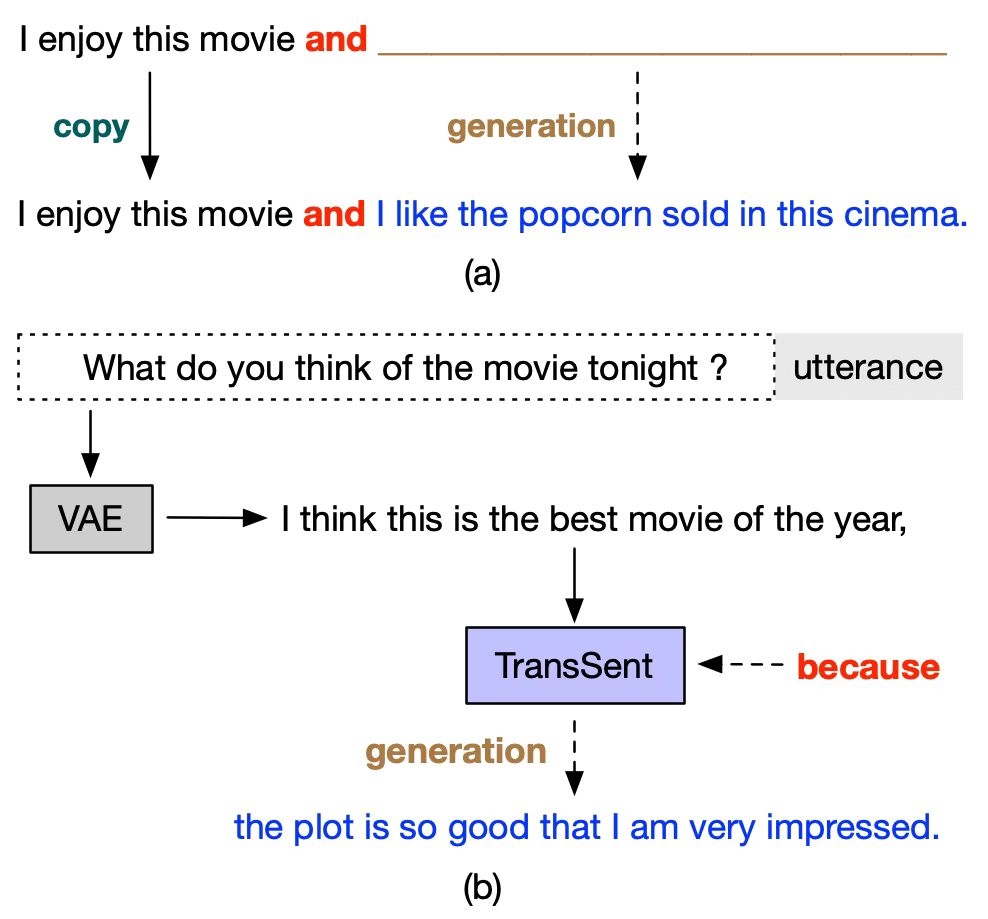}
	\caption{(a) An example of discourse transfer. (b) An application of discourse transfer in dialogue generation task.
} \label{trans_examples}
	\end{centering}
\end{figure}

Inspired by the ideas above, we focus on generating structured sentences with discourse markers.
We define a structured sentence as $(head$ $discourse,$ $relation,$ $tail$ $discourse)$, where the $relation$ is explicitly indicated by a discourse marker. We then propose a new task called \textbf{discourse transfer}, 
which aims at tail discourse generation based on head discourse and discourse marker. For instance, as shown in Figure \ref{trans_examples}-(a),
we want to generate a tail discourse that holds ``and" relation with the head discourse ``I enjoy the movie", such as "I like the popcorn sold in this cinema".
This setup brings two benefits.
\textbf{First}, the sentences generated by the proposed method are naturally structured with discourse markers.
\textbf{Second}, as humans naturally use discourse markers in writings or conversations, we can easily collect large amount of samples without hand annotation~\cite{DisSent}. 
In this paper, we construct three datasets (i.e., yelp-dm, wiki-dm and book-dm) from different sources, for model training and evaluation.

To fulfill the discourse transfer task, we propose a novel model called TransSent.
The backbone network of TransSent is an encoder-decoder network.
The encoder aims to learn the hidden representation of an input head discourse, while the decoder decodes the hidden representation back into the input. Given a relation, the decoder also learns to generate a tail discourse based on the head discourse and the corresponding relation.
As the hidden spaces of discourses and relations are not necessarily identical, we add an extra relation translation network, which is very similar to TransR~\cite{lin2015learning} in principle. 
TransR models entities and relations in distinct spaces, i.e., entity and relation spaces, and performs translation in
the relation space. In this paper,
the relation translation network also translates the discourse representation into a new hidden representation in the relation space. The decoder then decodes the new representation into a tail discourse.
We use BERT~\cite{BERT} as the encoder and unidirectional long-short term memory (LSTM) network as the decoder~\footnote{Considering that the previous related work used the RNN decoder, for the sake of fairness, we also choose the RNN decoder. And the beam search strategy is \textbf{not} used in the proposed model.}.
The relation translation network is a feed-forward network with nolinear activation. 
We experiment on three tasks: discourse transfer, free text generation and dialogue generation. Basically, we first evaluate how well TransSennt solve the discourse transfer task defined in our paper. Furthermore, to explore 
the scalability of TransSent, we combine TransSent with existing generative models, such as variational autoEncoder (VAE), to experiment in free text generation task.
Finally, by combining with the conditional VAE  (CAVE)~\cite{zhao2017learning}, TransSent is applied to the dialogue scenario, as shown in Figure~\ref{trans_examples}-(b).

In automatic evaluation, we use a language model as discriminator~\cite{yang2018unsupervised} to measure the coherence and cohesion of structured sentences. In addition, the discriminator of discourse marker prediction (DMP)~\cite{DisSent} task is used to measure the correctness of discourse relations.
For a more comprehensive evaluation of the quality of generated structured sentences, we also use human evaluation. 
The main contributions are summarized as follows:
\begin{itemize}
  \item We define the discourse transfer task and construct three datasets for further research.
  \item We propose a novel structured sentence generation model, TransSent, which has certain scalability in different tasks.
  \item Automatic and human evaluation results show that TransSent can generate structured sentences with high quality.
\end{itemize}

\section{Related Work}
\subsection{Long and Diverse Text Generation}
Natural language generation has drawn enough attention and been studied by many researchers.
\cite{bowman2015generating} uses VAE to generate sentences from continuous space.
\cite{yu2017seqgan} models text generation as a sequential decision making process by training the generative model with policy gradient strategy~\cite{sutton2000policy}.
\cite{zhao2017learning} also learns discourse-level information to generate diverse responses for neural dialogue tasks.
\cite{guo2018long} proposes LeakGAN for generating long text via adversarial training which allows the discriminative net to leak its own high-level extracted features to the generative net for further guidance.
\cite{bosselut2018discourse} investigates the reinforcement learning with discourse-aware rewards to guide a model in long text generation.
\cite{xiong2018modeling} proposes to use discourse context and reward for refining the translation quality from the discourse perspective.
\cite{shen2019towards} leverages several multi-level structures to train a VAE model for generating long and coherent text.
\cite{shao2019long} presents the planning-based hierarchical variational model (PHVM) for long and diverse text generation.
However, these works encode semantic information and structural information directly in one mixed hidden representation, making it hard to decode a well-structured and semantically meaningful sentence with sentence getting long and complicated. Instead, our model can naturally generate structured sentences with explicit discourse markers. Moreover, our model can also be combined with existing good methods like CVAE~\cite{zhao2017learning}.

\subsection{Translation Models for Knowledge Representation Learning}
Knowledge representation learning aims to embed the entities and relations in knowledge graphs (KGs) into a continuous semantic space. 
A knowledge graph is a set of fact triples with the format $(h, r, t)$, 
where $h$ and $t$ are the head and tail entities holding the relation $r$. 
TransE~\cite{bordes2013translating} attempts to regard a relation $r$ as a translation 
between the head and tail entities. It assumes that $h + r \approx t$ when $(h, r, t)$ holds. 
To address the issue of TransE when modeling complex mapping relations, 
TransH~\cite{wang2014knowledge} is proposed to enable an entity to have different representations when involved in various relations. 
TransR~\cite{lin2015learning} observes that an entity might exhibit different attributes in distinct relations, and models entities as well as relations in separated spaces.

\begin{figure}[htbp]
	\begin{centering}
	\includegraphics[width=0.45\textwidth]{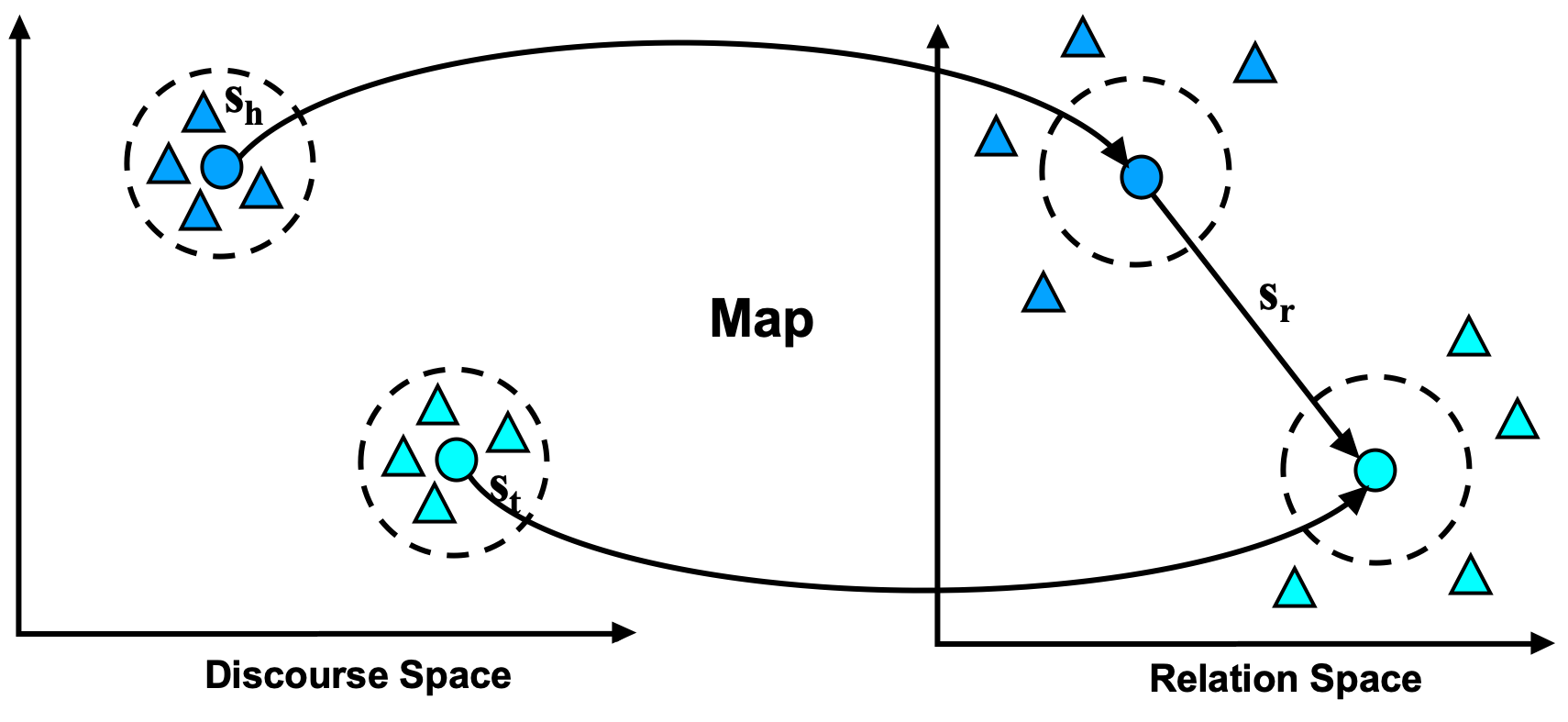}
	\caption{Simple illustration of TransSent.
} \label{trans method}
	\end{centering}
\end{figure}

Inspired by the success of these translation models, 
we perform the similar translation between two discourses in the embedding space, 
in order to simulate the semantic relationship indicated by a discourse marker, as shown in Figure \ref{trans method}.
One challenge is that sentences are more complicated than knowledge entities, when encoded into a continuous semantic space.
The other challenge is that the translated hidden representation needs to be able to decoded into text, which increases the difficulty of the problem.


\section{Discourse Transfer Task}
We focus here on structured sentences with explicit discourse markers between adjacent discourses, rather than implicit relation between a sentence and the related discourse.
We define a structured sentence as $(head$ $discourse,$ $relation,$ $tail$ $discourse)$, where the $relation$ is indicated by discourse markers.
Discourse transfer task aims at tail discourse generation based on head discourse and discourse marker. The generated tail discourse should hold the relation to the head discourse.
This task is a natural helper for generating structured sentences.
We can generate a head discourse with some existing methods such as VAE and choose a discourse marker. Given the head discourse and the discourse marker, we can then generate a tail discourse through discourse transfer. By concatenating these three parts, we will get an entire structured sentence. 



\section{TransSent Model}
In this section, we introduce TransSent, a novel model for discourse transfer, in details. It consists of two parts: an encoder-decoder and a relation translation network. The encoder learns the hidden representation of a discourse, while the decoder interprets the representation back to its original text. 
Relation translation network learns to translate a head discourse to its corresponding tail discourse 
according to a specified relation. 
Formally, given a training set of discourse pairs and relations between them, each example is denoted as $(\text{s}_h,\text{s}_r,\text{s}_t)$\footnote{Subscript $h$ represents $head$, $r$ represents $relation$ and $t$ represents $tail$.}, composed of head discourse $\text{s}_h$, tail discourse $\text{s}_t$, and a relation $\text{s}_r$.
Our model learns the embeddings of the discourses and the relations. 
As discourses and relations are different, it may be not capable to represent them in a common embedding space. 
We thus propose to model discourses and relations in distinct spaces, i.e. \textbf{discourse space} and \textbf{relation space}, then perform discourse transfer in relation space.

\begin{figure}[tbp]
	\begin{centering}
	\includegraphics[width=0.4\textwidth]{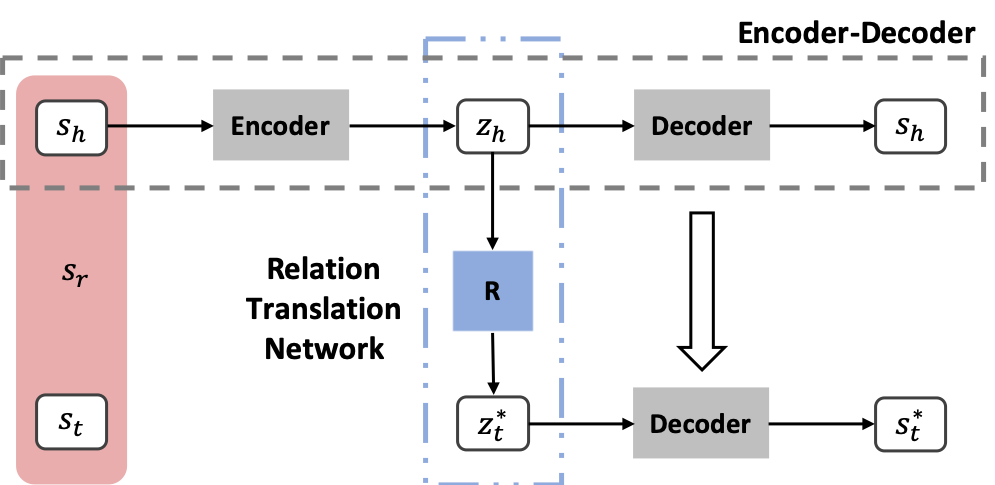}
  \caption{The overall model architecture of TransSent.} \label{model_arch}
	\end{centering}
\end{figure}

\paragraph{TransSent} denoted by $TransSent=(G,R)$, consists of two parts: an encoder-decoder network $G$ and 
a relation translation network $R$, as shown in Figure~\ref{model_arch}.
The encoder-decoder network $G$ comprises an encoder $G_{enc}$ and a decoder $G_{dec}$, denoted as $G=(G_{enc}, G_{dec})$.
$G_{enc}$ encodes the head discourse $\text{s}_h$ into its feature representation $\text{z}_h$, then the relation translation network $R$ exerts relation translation on $\text{z}_h$ and gets $\text{z}_t^*$. $\text{z}_t^*$ is the feature representation of the generated tail discourse. We expect $\text{z}_t^*$ to be as close as $\text{z}_t$ when training.
Finally, $G_{dec}$ decodes $\text{z}_t^*$ into a generated tail discourse $\text{s}_t^*$. We expect $\text{s}_t^*$ to hold the relation $\text{s}_r$ with $\text{s}_h$.

\subsection{Encoder-Decoder Network}
We use a pre-trained deep bidirectional transformer, known as BERT~\cite{BERT}, as our encoder $G_{enc}$. 
This model has proved to be able to provide good sentence representation.
To get a highly self-interpretable representation $z$ for a sentence or discourse, we fine-tune BERT on the discourse marker prediction (DMP) task~\cite{DisSent}. DMP task aims to predict which discourse marker should be taken to connect the two adjacent discourses. BERT takes a pair of adjacent discourses as input, and outputs the probability distribution on discourse markers. Each of the discourses is encoded in one input segment. We leave out the details of fine-tuning as there is an official open tutorial\footnote{\url{https://github.com/google-research/bert}} about it.
The fine-tune process can also be considered as teaching BERT to understand the structural relations between discourse pairs. One extra benefit is that the fine-tuned BERT is also used for automatic evaluation, to judge whether the discourse marker within a structured sentences is correct or not. 

We optimize the $G_{enc}$ by minimizing the following loss:
\begin{equation}
  {\cal{L}}_{enc} = -log p(r|s_h, s_t) \tag{1} \label{enc_loss}
\end{equation}
Then we keep the weights of fine-tuned BERT model fixed, and use it to extract the representations
of head discourse $s_h$ and tail discourse $s_t$, respectively.
Take $s_h$ for instance, supposing $s_h = \{s_{h_1},s_{h_2},...,s_{h_n}\}$, where $n$ is the number of total tokens, the fine-tuned BERT outputs hidden representations for each position in last layer, denoted as $o_h = \{o_{h_1},o_{h_2},...,o_{h_n}\}$. We concatenate each $o_{h_i}$ and apply nolinear projection, i.e., relu, on concatenated vector to get the head discourse's representation $z_h$:
\begin{equation}
    z_h = nolinear(W [o_{h_1} \oplus o_{h_2}\oplus...\oplus o_{h_n}]) \tag{2}
\end{equation}
where $\oplus$ denotes concatenation operation and $W$ is a projection matrix. 
We use the same method to extract the tail discourse's representation $z_t$.

\paragraph{Decoder}
We adopt unidirectional LSTM~\cite{LSTM} as the decoder $G_{dec}$ to recover $s_h$ and $s_t$ from $z_h$ and $z_t$, respectively. Take $z_h$ for instance:
\begin{align*}
  d &= {LSTM}(z_h)  \\
  p(s_t) &= {softmax}(d W) \tag{3}
\end{align*}
where $W$ is a projection matrix. 
We optimize the decoder by minimizing the following loss:
\begin{equation}
  {\cal{L}}_{rec} = -\frac{1}{2}(log p(s_h|z_h) + log p(s_t|z_t)) \tag{4} \label{rec_loss}
\end{equation}


\subsection{Relation Translation Network}
Following the similar setting in knowledge representation, 
we represent relation as translation in the embedding space.
We propose a hypothesis, under which we construct a relation translation network.

\paragraph{Hypothesis} Referring to TransE~\cite{bordes2013translating}, we assume:
\begin{equation}
  f(z_h,z_r) \approx z_t \tag{5} \label{hypothesis-2}
\end{equation}
That is, there exists a mapping in embedding space between a tail discourse $z_t$ and a head discourse $z_h$ plus some vector that depends on the relation representation $z_r$. 
Formally, in relation translation network, each triple is denoted as $(z_h, z_r, z_t)$, where head discourse and tail discourse are represented as $z_h, z_t \in R^k$ and relation is represented as $z_r \in R^d$.
Referring to TransR~\cite{lin2015learning}, the dimension of sentence and relation embeddings are not necessarily identical, i.e., $k \neq d$.
For each relation $s_r$, we train a projection matrix $M_r \in R^{k \times d}$ , which projects discourse representations from discourse space to relation space. With the projection matrix, we then define the projected representation vectors of sentences as:
\begin{equation}
{z_h}_r = {z_hM}_r, 
{z_t}_r={z_tM}_r \tag{6}  \label{translation-2}
\end{equation}
Specifically, in relation space, for $({z_h}_r, r, {z_t}_r)$, we apply translation on ${z_h}_r$: 
\begin{equation}
{z_t}_r^* = W^{\prime}[{z_h}_r \oplus z_r] \tag{7}
\end{equation}
where $W^{\prime}$ is a projection matrix. We then measure the distance between the two vectors in relation space with L2-norm distance:
\begin{equation}
  L_{dis}=||{z_t}_r^* - {z_t}_r||_2^2 \tag{8} \label{dis_loss}
\end{equation} 
To further encourage ${z_t}_r^*$ to be as close as ${z_t}_r$, we want ${z_t}_r^*$ is much farther from ${z_h}_r$, than from ${z_t}_r$. So we introduce another loss:
\begin{equation}
  L_{ratio}=||W[{z_h}_r \oplus z_r] - {z_t}_r||_2^2 / ||W[{z_h}_r \oplus z_r] - {z_h}_r||_2^2 
  \tag{9}
  \label{ratio_loss}
\end{equation}
We feed ${z_t}_r^*$ into a feed-forward network to map ${z_t}_r^*$ from relation space back into discourse space ${z_t}^*$, which approximates matrix inversion operation $M^{-1}$ with neural computation.
Then $s_t^*$ can be obtained through decoder with Equation 3, i.e., $s_t^* = G_{dec}({z_t}^*)$.
Ideally, $s_t^*$ holds the original relation $s_r$ with original head discourse $s_h$. 
Combing Equation~\ref{rec_loss}~\ref{dis_loss}~\ref{ratio_loss}
, we can obtain the training objective:
\begin{equation}
  \min_{\theta}{\cal{L}} = {\cal{L}}_{rec} + \lambda_{d}{\cal{L}}_{dis} + \lambda_{r}{\cal{L}}_{ratio} \tag{10} \label{loss}
\end{equation} 
where $\lambda_{d}$ and $\lambda_{r}$ are balancing hyperparameters.
The training details of TransSent are shown in Algorithm \ref{alg1}.
\begin{algorithm}[h]
  \caption{Implementation of TransSent model in details }\label{alg1}
  \begin{algorithmic}[1]
  \STATE Fine-tune BERT with DMP task and get $G_{enc}$
  \STATE Fix the weights of $G_{enc}$
  \FOR {each iteration i=1,2,...,M}
    \STATE Sample a structured sentence $\{s_h,s_r,s_t\}$ 
    \STATE Obtain discourse representations $z_h$ and $z_t$ based on Eq.2
    \STATE Calculate reconstruction loss ${\mathcal{L}_{rec}}$ with $G_{dec}$ based on Eq.4
    \STATE Do relation translation, acquire new tail discourse representation $z_t^*$ based on Eq.6-7
    \STATE Calculate ${\mathcal{L}_{dis}}$ and ${\mathcal{L}_{ratio}}$ based on Eq.8-9
    \STATE Calculate ${\mathcal{L}}$ based on Eq.10
  	\STATE Update model parameters $\theta$
  \ENDFOR
  \end{algorithmic}
\end{algorithm}


After well-trained on discourse transfer task, our TransSent can be used to generate structured sentences combined with existing models. Take free text generation task for instance,
we first adopt well-trained VAE to randomly generate a head discourse, 
then we select a discourse marker and perform discourse transfer with TransSent, generating the tail discourse. 
At last, we concatenate the head discourse and the tail discourse with the discourse marker to get an entire structured sentence.


\section{Data Collection}
We use public code for data collection from~\cite{DisSent}~\footnote{\url{https://github.com/windweller/DisExtract}}.
There are many discourse markers and we focus on the set of five most common ones, i.e., ~\textbf{and}, \textbf{but}, \textbf{because}, \textbf{if} and \textbf{when}.
We collect an in-domain and two open-domain datasets from natural text corpora using explicit discourse markers. Each dataset consists of discourse pairs and the relations between them. We randomly split each dataset into training, development and test sets.
\begin{itemize}
\item\textbf{Yelp-dm}($10$K/$1$K/$1$K) is extracted from a sentiment domain corpus Yelp, 
which consists of movie reviews from business review website $Yelp$\footnote{\url{http://www.yelp.com}}.
\item\textbf{Wiki-dm}($80$K/$5$K/$3$K) is extracted from an open-domain corpus WikiText-103\cite{merity2016pointer}, which consists of Wikipedia articles. 
\item\textbf{Book-dm}($400$K/$30$K/$30$K) is extracted from another open-domain corpus BookCorpus~\cite{zhu2015aligning}, which consists of texts from unpublished novels.
\end{itemize}

\paragraph{Dependency Parsing} The positions of discourse markers relative to their connected discourses can vary. For example, ``Because [it was cold outside]$_{S2}$, [I wore a jacket]$_{S1}$" equals to ``[I wore a jacket]$_{S1}$, because [it was cold outside]$_{S2}$". So following~\cite{DisSent}, we use Stanford CoreNLP dependency parser~\cite{schuster2016enhanced} to extract the appropriate pairs of sentences and filter them based on the order of sentences in the original text. Furthermore,
we exclude any cases where one of the two discourses is less than 5 or more than 15 words.
We randomly select the same amount of pairs for each discourse marker and add them into corpus.

\section{Experiments}
We first evaluate TransSent's ability to generate coherent and structured sentences in discourse transfer task.
To explore how our model can be combined  with  existing  generative  models, such as VAE, we further experiment on free text generation task. We replace the collected head discourses in our datasets with generated head discourses by VAE. We also apply our model to dialogue generation task with CVAE. Same as VAE, CVAE is used to generate head responses that are conditional on historical utterances, as shown in figure \ref{trans_examples}-(b).

\subsection{Baselines}
\subsubsection{Discourse Transfer Task}
We compared TransSent with sequence-to-sequence-based and plan-based (PHVM) models on free text generation task.
\paragraph{Seq2Seq}~\cite{nguyen2017sequence}: A sequence-to-sequence learning method which generate sentences describing a predicted event from the preceding sentences based on a bidirectional multi-layer recurrent neural network (RNN).
\paragraph{PHVM}~\cite{shao2019long}: A plan-based model that first generates a sequence of groups, each of which is a subset of attribute-value pairs. Then, each sentence is realized from its group and previous sentences based on CVAE. We adapt it to our task by replacing the attribute-value pairs with discourse markers and fixing the initial sentences to head discourses. 
\subsubsection{Free Text Generation Task}
We compared TransSent with sequence-to-sequence (mentioned before), VAE, and GAN-based (LeakGAN) models on free text generation task.
\paragraph{VAE}~\cite{bowman2015generating}: A RNN-based variational autoencoder model which incorporates distributed hidden representations of entire sentences for text generation. 
\paragraph{LeakGAN}~\cite{guo2018long}: LeakGAN designs a discriminative net to leak its own high-level extracted features to the generative net. Given the features, generative net then utilizes a Manager-Worker framework to learn sentence structures and generate long sentences. 
\subsubsection{Dialogue Generation Task} We compared TransSent with CVAE and plan-based (PHVM) models (mentioned before) on dialogue generation task.
\paragraph{CVAE}~\cite{zhao2017learning} utilizes latent variables to learn a distribution over dialogue acts, such as \textit{statement} and \textit{yes-answer}, then generates diverse responses based on RNN decoders.

\subsection{Experiment Details}
For TransSent,
we fine-tune BERT$_{base}$ with public model configuration that can be found in \cite{BERT} in the DMP task for $8$ epochs.
The fine-tuned Bert achieves accuracy of $87.1\%$, $86.7\%$, $88.8\%$ on Yelp-dm, Wiki-dm and Book-dm on DMP task, respectively. 
We fix the weights of the fine-tuned BERT and use the fine-tuned Bert as our encoder and discriminator. We concatenate the token representations of discourses from the top layers of the encoder and use an affine fully-connected layer to project the concatenated vector to the discourse representation. The relation network then projects the head discourse representation into relation space and concatenates it with relation vector, then feeds the concatenated vector into a feed-forward network to obtain the representation of tail discourse.
We adopt a single layer unidirectional LSTM network to decode the tail representation into a tail discourse, with dropout set to $0.1$. We train our model for $20$ epochs, with Adam \cite{kingma2014adam} as our optimizer. The embedding dimensions are all set to $768$. All baselines are implemented according to relevant papers and public codes. In particular, in dialogue generation task, we first fine-tune the model on Book-dm dataset, and continue to fine-tune the model on training data provided by CAVE~\cite{zhao2017learning}.

\subsection{Evaluation}

\begin{table*} [!h]
\small
\centering
\caption{Examples of free text generation by TransSent.}\label{free_text_gen}
\begin{tabular}{l|c}
  \toprule
   and & They were initially unhappy, \textbf{and} were surprised when fans responded positively to it. \\
   but & I saw glimpses of it when I reviewed her file \textbf{but} when I met her face-to-face , i knew it was true. \\
   because & I only gave it one star, \textbf{because} this restaurant wasn't for the lunch. \\
   if & How could she resist him \textbf{if} he had this power over her and made a move to claim her.\\
   when & Then , in late fall i moved back to copper creek \textbf{when} it 's dark early and begins to rain in earnest. \\
  \bottomrule
  \end{tabular}
\end{table*}

\begin{table*}[!h]
  \small
  \centering
  \caption{Evaluation results on discourse transfer and free text generation tasks. ACC, NLL and PPL are automatic evaluation metrics; Gra(grammaticality) and Coh(coherence) are human evaluation ratings. For NLL and PPL lower is better, while for others higher is better.}\label{tab3}
  \setlength{\tabcolsep}{1.9mm}
  \begin{tabular}{l|ccccc|ccccc|ccccc}
  \toprule
        & \multicolumn{5}{c|}{Yelp-dm} & \multicolumn{5}{c|}{Wiki-dm} & \multicolumn{5}{c}{Book-dm}\\
        &ACC & PPL & NLL &Gra & Coh &ACC & PPL & NLL &Gra & Coh & ACC & PPL & NLL &Gra & Coh\\
  \midrule
  \textit{\textbf{discourse transfer}} & & & & & & & & & & & & & & & \\
  Seq2Seq & 19.7& 137.2& 4.9& 2.1& 2.2& 15.6& 151.9& 5.0& 1.9& 2.0& 12.3& 208.8& 5.3& 2.0& 2.1 \\
  PHVM  & 27.4& 88.0 & 4.5& 2.9& 2.8& 23.5&119.9& 4.7& 2.8&  2.9& 23.2&100.3& 4.9& 2.4& 2.5 \\
  TransSent  & \textbf{79.8}&  \textbf{65.4}& \textbf{4.2}& \textbf{3.2}& \textbf{3.1}& \textbf{57.8}&\textbf{87.5}& \textbf{4.7}& \textbf{3.3}& \textbf{3.2}& \textbf{51.2}& \textbf{63.4}& \textbf{4.2}& \textbf{3.2}& \textbf{3.1} \\
  \midrule
  \textit{\textbf{free text generation}} & & &  & & & & & &  & & & & & & \\
  Seq2Seq & 10.3& 154.3& 5.0& 1.8& 1.9& 9.7& 155.4& 5.0& 1.7& 1.8& 14.7& 200.4& 5.3& 1.6&  1.7\\
  VAE & 11.4&  118.1&  4.7& 2.2& 2.3&  10.5& 158.9& 5.0&  2.1& 2.0& 13.6& 160.2& 5.1& 2.2& 2.0\\
  LeakGAN & 27.2&  114.0&  4.7& 2.8& 2.8& 25.3& 133.0& 4.9&  2.5& 2.6& 22.7& 94.9& 4.6& 2.4& 2.3\\
  TransSent&  \textbf{67.0}& \textbf{100.9}&  \textbf{4.6}& \textbf{3.0}& \textbf{3.1}& \textbf{53.3}& \textbf{90.2}& \textbf{4.5}&  \textbf{3.1}& \textbf{3.2}& \textbf{47.4}& \textbf{88.5}& \textbf{4.5}& \textbf{3.0}& \textbf{3.1}\\
  \bottomrule
  \end{tabular}
  \end{table*}

 \begin{table}[!h]
  \small
  \centering
  \caption{Evaluation results on dialogue generation task.}\label{tab4}
  \begin{tabular}{l|ccccc}
  \toprule
        &ACC & PPL & NLL &Gra & Coh \\
  \midrule
  CVAE & 17.4& 176.2 & 4.8 & 2.5 & 2.7  \\
  PHVM & 23.5& 104.1 & 4.6 & 3.0 & 2.9 \\
  TransSent & \textbf{54.2}& \textbf{90.7} &\textbf{4.5} & \textbf{3.2} & \textbf{3.4}\\
  \bottomrule
  \end{tabular}
  \end{table}
  
\begin{table} [!h]
\small
\centering
\caption{Examples of dialogue response generated by different models.}\label{dialog_gen}
\begin{tabular}{l}
  \toprule
    $1.$
	\textcolor{blue}{Well I think air pollution is a pretty big problem right now \textbf{and}} \\ 
    CVAE: I don't think that’s a lot.\\
    PBHM: is really being done about it at this point.\\
    TransSent: \textcolor{red}{I think it's a real bad problem.}\\
	\midrule
	$2.$
    \textcolor{blue}{Personal computers are nice I guess, \textbf{if}} \\
    CVAE: you happen at work.\\
    PBHM: you tend to get spoiled.\\
    TransSent: \textcolor{red}{you can afford them.}\\
  \bottomrule
  \end{tabular}
\end{table}

\subsubsection{Automatic Evaluation}
\paragraph{Accuracy}
We use the fine-tuned BERT as the discriminator to assess whether relation between discourses in generated sentences holds.
\paragraph{Coherence and Cohesion}
To evaluate the coherence and cohesion of different models, we report the scores of two widely used metrics, negative log-likelihood (NLL) and perplexity (PPL).~\footnote{As our sentences are generated from scratch randomly, without any references, so BLEU metric fits not here.} The perplexity loss indicates the continuity and coherence of content between tail discourse and head discourse. 

\subsubsection{Human Evaluation}
We hired five annotators to rate the outputs of TransSent and baselines. We adopt two criteria range from $1$ to $5$ ($1$ is very bad and $5$ is very good): grammaticality, relation correctness to the target attribute. For each model on each dataset, we randomly sample $200$ generated examples. Noting that all the comparision are done on the generated tails with same head discourses and discourse markers.

\subsection{Results Analysis}
Through automatic evaluation and human evaluation, as shown in Table~\ref{tab3} and Table~\ref{tab4}, our model outperforms baselines in all respects. Our model achieves very higher ACC on all datasets, which benefits from that we explicitly separate the modeling process of semantic information and structural information. Since discourses are shorter than an entire sentence, the long-term dependencies problem during decoding is relieved. As a result, our model is superior to baselines in both PPL and NLL.

In addition, better results in free text generation task show that our model can be easily and effectively combined with existing generation models (such as VAE). We show some generated samples in Table~\ref{free_text_gen}, and our model is able to generate well-structured and meaningful sentences.

Lastly, our model also achieves satisfactory performance in dialogue scenario. We also show two cases in Table~\ref{dialog_gen}.

\subsection{Deficiencies}
As shown in Table~\ref{tab3}, the relation accuracies on Wiki-dm and Book-dm datasets are not as good as on Yelp-dm. This is because the samples collected in Yelp-dm are all movie reviews, most of which share structural and semantic similarities. In contrast, Wiki-dm and Book-dm are from open-domain sources. They are more general and diverse, which makes it difficult for models to better learn relation translation networks. Therefore, it is still a challenge to explore a better model architecture or collect a larger dataset to obtain satisfactory results in open-domain structured sentence generation.

\section{Conclusions}
In this paper, we focus on generating long sentences with explicit structure. To achieve this, we define a new task \textbf{discourse transfer}, which generates tail discourse based on head discourse and discourse marker, and constructs three datasets from different domains for this task. We then propose a novel model called TransSent, which translates the representation of a head discourse to a tail discourse in the relation embedding space and outputs a structured sentence through decoding and concatenating. 
The experimental results through automatic evaluation and human evaluation show that our method can generate structured sentences with higher quality.


\section{Future Work}
There are more than the five most common discourse markers we focus on in this paper, so in the future we will explore more discourse markers and construct larger corpus for further research.
Also, we will focus on generating long structured sentences with several discourse markers in a recursive way.

\newpage
\newpage
\bibliographystyle{named}
\bibliography{ijcai20}

\begin{thebibliography}{}

\bibitem[\protect\citeauthoryear{Bordes \bgroup \em et al.\egroup
  }{2013}]{bordes2013translating}
Antoine Bordes, Nicolas Usunier, Alberto Garcia-Duran, Jason Weston, and Oksana
  Yakhnenko.
\newblock Translating embeddings for modeling multi-relational data.
\newblock In {\em Advances in neural information processing systems}, pages
  2787--2795, 2013.

\bibitem[\protect\citeauthoryear{Bosselut \bgroup \em et al.\egroup
  }{2018}]{bosselut2018discourse}
Antoine Bosselut, Asli Celikyilmaz, Xiaodong He, Jianfeng Gao, Po-Sen Huang,
  and Yejin Choi.
\newblock Discourse-aware neural rewards for coherent text generation.
\newblock {\em arXiv preprint arXiv:1805.03766}, 2018.

\bibitem[\protect\citeauthoryear{Bowman \bgroup \em et al.\egroup
  }{2015}]{bowman2015generating}
Samuel~R Bowman, Luke Vilnis, Oriol Vinyals, Andrew~M Dai, Rafal Jozefowicz,
  and Samy Bengio.
\newblock Generating sentences from a continuous space.
\newblock {\em arXiv preprint arXiv:1511.06349}, 2015.

\bibitem[\protect\citeauthoryear{Devlin \bgroup \em et al.\egroup
  }{2018}]{BERT}
Jacob Devlin, Ming{-}Wei Chang, Kenton Lee, and Kristina Toutanova.
\newblock {BERT:} pre-training of deep bidirectional transformers for language
  understanding.
\newblock {\em CoRR}, abs/1810.04805, 2018.

\bibitem[\protect\citeauthoryear{Guo \bgroup \em et al.\egroup
  }{2018}]{guo2018long}
Jiaxian Guo, Sidi Lu, Han Cai, Weinan Zhang, Yong Yu, and Jun Wang.
\newblock Long text generation via adversarial training with leaked
  information.
\newblock In {\em Thirty-Second AAAI Conference on Artificial Intelligence},
  2018.

\bibitem[\protect\citeauthoryear{Hobbs}{1990}]{hobbs1990literature}
Jerry~R Hobbs.
\newblock Literature and cognition.
\newblock (21), 1990.

\bibitem[\protect\citeauthoryear{Hochreiter and Schmidhuber}{1997}]{LSTM}
Sepp Hochreiter and J{\"{u}}rgen Schmidhuber.
\newblock Long short-term memory.
\newblock {\em Neural Computation}, 9(8):1735--1780, 1997.

\bibitem[\protect\citeauthoryear{Hu \bgroup \em et al.\egroup
  }{2017}]{hu2017toward}
Zhiting Hu, Zichao Yang, Xiaodan Liang, Ruslan Salakhutdinov, and Eric~P Xing.
\newblock Toward controlled generation of text.
\newblock In {\em Proceedings of the 34th International Conference on Machine
  Learning-Volume 70}, pages 1587--1596. JMLR. org, 2017.

\bibitem[\protect\citeauthoryear{Ilya~Sutskever and
  Le}{2014}]{Sutskever-2014-NIPS}
Oriol~Vinyals Ilya~Sutskever and Quoc~V Le.
\newblock Sequence to sequence learning with neural net- works.
\newblock {\em In Advances in neural information process- ing systems}, 2014.

\bibitem[\protect\citeauthoryear{Kikuchi \bgroup \em et al.\egroup
  }{2016}]{kikuchi2016controlling}
Yuta Kikuchi, Graham Neubig, Ryohei Sasano, Hiroya Takamura, and Manabu
  Okumura.
\newblock Controlling output length in neural encoder-decoders.
\newblock {\em arXiv preprint arXiv:1609.09552}, 2016.

\bibitem[\protect\citeauthoryear{Kingma and Ba}{2014}]{kingma2014adam}
Diederik~P Kingma and Jimmy Ba.
\newblock Adam: A method for stochastic optimization.
\newblock {\em arXiv preprint arXiv:1412.6980}, 2014.

\bibitem[\protect\citeauthoryear{Lin \bgroup \em et al.\egroup
  }{2015}]{lin2015learning}
Yankai Lin, Zhiyuan Liu, Maosong Sun, Yang Liu, and Xuan Zhu.
\newblock Learning entity and relation embeddings for knowledge graph
  completion.
\newblock In {\em Twenty-ninth AAAI conference on artificial intelligence},
  2015.

\bibitem[\protect\citeauthoryear{Merity \bgroup \em et al.\egroup
  }{2016}]{merity2016pointer}
Stephen Merity, Caiming Xiong, James Bradbury, and Richard Socher.
\newblock Pointer sentinel mixture models.
\newblock {\em arXiv preprint arXiv:1609.07843}, 2016.

\bibitem[\protect\citeauthoryear{Nguyen \bgroup \em et al.\egroup
  }{2017}]{nguyen2017sequence}
Dai~Quoc Nguyen, Dat~Quoc Nguyen, Cuong~Xuan Chu, Stefan Thater, and Manfred
  Pinkal.
\newblock Sequence to sequence learning for event prediction.
\newblock {\em arXiv preprint arXiv:1709.06033}, 2017.

\bibitem[\protect\citeauthoryear{Nie \bgroup \em et al.\egroup
  }{2017}]{DisSent}
Allen Nie, Erin~D. Bennett, and Noah~D. Goodman.
\newblock Dissent: Sentence representation learning from explicit discourse
  relations.
\newblock {\em CoRR}, abs/1710.04334, 2017.

\bibitem[\protect\citeauthoryear{Rajeswar \bgroup \em et al.\egroup
  }{2017}]{rajeswar2017adversarial}
Sai Rajeswar, Sandeep Subramanian, Francis Dutil, Christopher Pal, and Aaron
  Courville.
\newblock Adversarial generation of natural language.
\newblock {\em arXiv preprint arXiv:1705.10929}, 2017.

\bibitem[\protect\citeauthoryear{Schuster and
  Manning}{2016}]{schuster2016enhanced}
Sebastian Schuster and Christopher~D Manning.
\newblock Enhanced english universal dependencies: An improved representation
  for natural language understanding tasks.
\newblock In {\em LREC}, pages 23--28. Portoro{\v{z}}, Slovenia, 2016.

\bibitem[\protect\citeauthoryear{Shao \bgroup \em et al.\egroup
  }{2019}]{shao2019long}
Zhihong Shao, Minlie Huang, Jiangtao Wen, Wenfei Xu, and Xiaoyan Zhu.
\newblock Long and diverse text generation with planning-based hierarchical
  variational model.
\newblock {\em arXiv preprint arXiv:1908.06605}, 2019.

\bibitem[\protect\citeauthoryear{Shen \bgroup \em et al.\egroup
  }{2019}]{shen2019towards}
Dinghan Shen, Asli Celikyilmaz, Yizhe Zhang, Liqun Chen, Xin Wang, Jianfeng
  Gao, and Lawrence Carin.
\newblock Towards generating long and coherent text with multi-level latent
  variable models.
\newblock {\em arXiv preprint arXiv:1902.00154}, 2019.

\bibitem[\protect\citeauthoryear{Sutton \bgroup \em et al.\egroup
  }{2000}]{sutton2000policy}
Richard~S Sutton, David~A McAllester, Satinder~P Singh, and Yishay Mansour.
\newblock Policy gradient methods for reinforcement learning with function
  approximation.
\newblock In {\em Advances in neural information processing systems}, pages
  1057--1063, 2000.

\bibitem[\protect\citeauthoryear{Wang \bgroup \em et al.\egroup
  }{2014}]{wang2014knowledge}
Zhen Wang, Jianwen Zhang, Jianlin Feng, and Zheng Chen.
\newblock Knowledge graph embedding by translating on hyperplanes.
\newblock In {\em Twenty-Eighth AAAI conference on artificial intelligence},
  2014.

\bibitem[\protect\citeauthoryear{Xiong \bgroup \em et al.\egroup
  }{2018}]{xiong2018modeling}
Hao Xiong, Zhongjun He, Hua Wu, and Haifeng Wang.
\newblock Modeling coherence for discourse neural machine translation.
\newblock {\em arXiv preprint arXiv:1811.05683}, 2018.

\bibitem[\protect\citeauthoryear{Yang \bgroup \em et al.\egroup
  }{2018}]{yang2018unsupervised}
Zichao Yang, Zhiting Hu, Chris Dyer, Eric~P Xing, and Taylor Berg-Kirkpatrick.
\newblock Unsupervised text style transfer using language models as
  discriminators.
\newblock In {\em Advances in Neural Information Processing Systems}, pages
  7287--7298, 2018.

\bibitem[\protect\citeauthoryear{Yu \bgroup \em et al.\egroup
  }{2017}]{yu2017seqgan}
Lantao Yu, Weinan Zhang, Jun Wang, and Yong Yu.
\newblock Seqgan: Sequence generative adversarial nets with policy gradient.
\newblock In {\em Thirty-First AAAI Conference on Artificial Intelligence},
  2017.

\bibitem[\protect\citeauthoryear{Zhao \bgroup \em et al.\egroup
  }{2017}]{zhao2017learning}
Tiancheng Zhao, Ran Zhao, and Maxine Eskenazi.
\newblock Learning discourse-level diversity for neural dialog models using
  conditional variational autoencoders.
\newblock {\em arXiv preprint arXiv:1703.10960}, 2017.

\bibitem[\protect\citeauthoryear{Zhu \bgroup \em et al.\egroup
  }{2015}]{zhu2015aligning}
Yukun Zhu, Ryan Kiros, Rich Zemel, Ruslan Salakhutdinov, Raquel Urtasun,
  Antonio Torralba, and Sanja Fidler.
\newblock Aligning books and movies: Towards story-like visual explanations by
  watching movies and reading books.
\newblock In {\em Proceedings of the IEEE international conference on computer
  vision}, pages 19--27, 2015.

\end{thebibliography}

\end{document}